\documentclass[runningheads]{llncs}

\usepackage{eccv}

\usepackage{eccvabbrv}
\usepackage{array}
\usepackage{subcaption}
\usepackage{graphicx}
\usepackage{booktabs}

\usepackage[accsupp]{axessibility}  

\usepackage{hyperref}

\usepackage{orcidlink}

\begin{document}

\title{AgentIAD: Agentic Industrial Anomaly Detection via Adaptive Memory Augmentation} 

\titlerunning{AgentIAD}

\author{
Junwen Miao\inst{1}\thanks{Equal contribution. $^\dagger$ Corresponding author} \and
Penghui Du\inst{1,2}$^\star$\and
Yingying Fan\inst{3}$^\star$\and
Yi Liu\inst{2} \and
Yu Wang\inst{4} \and \\
Runze He\inst{2} \and
Lida Huang\inst{1} \and
Yan Wang\inst{1}$^\dagger$
}

\authorrunning{
J. Miao, P. Du, Y. Fan~Author et al.}

\institute{
${^1}$AIR, Tsinghua University ~~${^2}$Etude AI ~~${^3}$Stony Brook University \\  ~~${^4}$Tsinghua University\\
\email{
miaojunwen9@gmail.com ~~wangyan@air.tsinghua.edu.cn
}}




\maketitle
\begin{abstract}
Industrial anomaly detection (IAD) is challenging due to the subtle and highly localized nature of many defects, which single-pass vision--language models (VLMs) often fail to capture. Moreover, existing approaches lack mechanisms to actively acquire complementary evidence during inference. We propose \textbf{AgentIAD}, an agentic vision--language framework that enables iterative industrial inspection through a unified action space. The agent dynamically accesses two forms of memory during inspection: \textbf{visual memory} via the \textbf{Perceptive Zoomer (PZ)} for fine-grained local analysis, and \textbf{retrieved memory} via the \textbf{Web Searcher (WS)} and \textbf{Comparative Retriever (CR)} for external knowledge acquisition and cross-instance verification. This design allows the model to progressively gather evidence through multi-round perception--action reasoning. To effectively learn such policies under sparse supervision, AgentIAD adopts a \textbf{two-stage training strategy}: tool-aware supervised fine-tuning first initializes structured reasoning and memory-access behaviors, followed by agentic reinforcement learning to refine long-horizon decision policies. Extensive experiments show that, under the same backbone, AgentIAD improves classification accuracy by \textbf{5.92\%} over the previous state-of-the-art method on the MMAD benchmark while providing more reliable and interpretable anomaly analysis.
\end{abstract}    
\section{Introduction}

Industrial anomaly detection (IAD) plays a vital role in automated manufacturing, ensuring product reliability, reducing waste, and preventing safety risks. Traditional embedding-based~\cite{bergmann2019mvtec,defard2021padim,pathcore} and reconstruction-based methods~\cite{zavrtanik2021draem,gong2019memorizing} have demonstrated strong performance in detecting structural and textural defects. However, industrial inspection inherently suffers from the scarcity of defective samples, requiring models to generalize reliably to previously unseen anomalies. Moreover, defects are often subtle and highly localized within complex industrial scenes, making fine-grained anomaly identification particularly challenging.

\begin{figure}[!t]
    \centering
    \includegraphics[width=0.95\columnwidth]{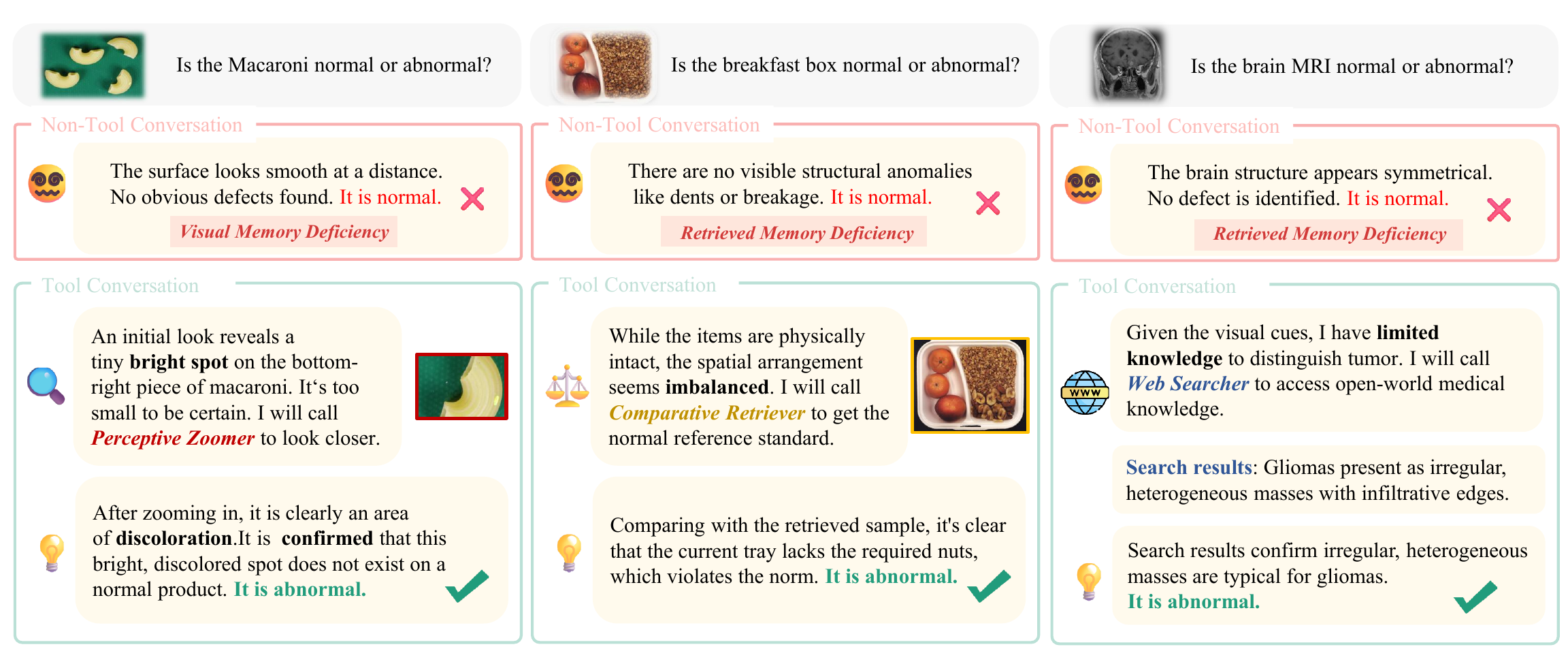}
    \caption{\textbf{Comparison with Non-tool MLLMs.}
Non-tool MLLMs rely on a single global pass and often miss subtle defects (left).
In contrast, AgentIAD performs iterative inspection through dynamic memory access (middle \& right): the \textit{Perceptive Zoomer} retrieves fine-grained \textit{visual memory} to reveal subtle anomalies, while the \textit{Comparative Retriever} and \textit{Web Searcher} access \textit{retrieved memory} to verify abnormal cues and obtain external knowledge for unfamiliar anomaly types. }
    \label{fig:motivation} 
\end{figure}

Recently, Multimodal Large Language Models (MLLMs)~\cite{chao2025anomalyr1,zeng2025lr,liao2025ad,guan2025emit,li2025iad,zhao2025omniad} have significantly advanced industrial anomaly detection (IAD) by enhancing semantic understanding and reasoning capabilities. However, most existing approaches still rely on a static single-pass inference paradigm, where the model processes a fixed image once and produces a final prediction without iterative perception or interaction with external knowledge sources. This design implicitly assumes that all relevant information required for anomaly analysis is available from the initial visual input and the model’s internal knowledge. As illustrated in Fig.~\ref{fig:motivation}, this assumption leads to two fundamental memory limitations: \textbf{1) Visual Memory Deficiency}, where subtle defects may remain unnoticed when only coarse visual representations are processed, and \textbf{2) Retrieved Memory Deficiency}, where models cannot access external references or domain knowledge needed to interpret unfamiliar anomalies. These limitations suggest that effective IAD requires models capable of dynamically accessing and integrating multiple forms of memory during inference.

We argue that inspiration for addressing these memory deficiencies can be drawn from human quality inspectors. In practice, expert inspectors rely on multiple forms of memory during inspection. They repeatedly examine suspicious regions to accumulate \textbf{fine-grained visual memory} of local details, and consult \textbf{external references or historical samples} when encountering unfamiliar patterns. These heterogeneous memories are accessed adaptively through an iterative \textit{Perception--Action Loop}, where each observation enriches the inspector’s understanding and guides subsequent actions. This behavior suggests a key insight: high-precision IAD is fundamentally not a static classification problem, but rather a sequential decision-making process that \textit{dynamically queries visual and retrieved memory}.

Building on this insight, we introduce \textbf{AgentIAD}, an agentic framework that performs industrial anomaly detection through iterative memory access within a perception--action loop. We formulate the detection process as a Markov Decision Process in which an agent learns a policy $\pi_\theta$ to coordinate perception and retrieval actions. To address the two memory deficiencies of static inference, AgentIAD enables two complementary forms of memory access: \textbf{1) Visual Memory}, accessed via the \textit{Perceptive Zoomer (PZ)}, which performs high-resolution inspection of suspicious regions to recover fine-grained visual details and mitigate the Visual Memory Deficiency; and \textbf{2) Retrieved Memory}, accessed via the \textit{Web Searcher (WS)} and \textit{Comparative Retriever (CR)}, which retrieve external knowledge and reference samples to overcome the Retrieved Memory Deficiency. 
Through iterative perception--action reasoning, the agent progressively localizes ambiguous regions, retrieves complementary evidence, and derives reliable and interpretable conclusions. 
However, effectively training such an agent remains challenging. Supervised fine-tuning on expert trajectories can provide initial tool-use capabilities but often results in policies with limited generalization to unseen scenarios. Reinforcement learning, on the other hand, suffers from inefficient exploration in the large combinatorial action space induced by multi-tool interactions, especially under sparse reward signals. To address this challenge, we adopt a pragmatic \textbf{two-stage training strategy}: we first perform \textit{Tool-Aware Supervised Fine-Tuning} to initialize the agent with structured tool-use behaviors, and then refine the policy through \textit{Agentic Reinforcement Learning}. This formulation improves training stability and generalization while enhancing interpretability by externalizing the reasoning process into explicit perception and tool-invocation steps.

Our main contributions are summarized as follows:
\begin{itemize}

\item We propose \textbf{AgentIAD}, a unified agentic framework for industrial anomaly detection that formulates the task as a Markov Decision Process and enables iterative perception–action reasoning for adaptive visual inspection.

\item We introduce a memory-augmented inspection paradigm that addresses two key limitations of static inference: \textbf{1) Visual Memory} for fine-grained perception via the Perceptive Zoomer to mitigate the resolution bottleneck, and \textbf{2) Retrieved Memory} for external knowledge acquisition and reference verification via the Web Searcher and Comparative Retriever.

\item We propose a pragmatic \textbf{two-stage training strategy} that combines Tool-Aware Supervised Fine-Tuning with Agentic Reinforcement Learning, enabling stable training of tool-augmented agents while improving generalization to complex inspection scenarios.

\item Extensive experiments show that AgentIAD achieves state-of-the-art performance on multiple IAD benchmarks, and our ablation studies validate the effectiveness of the proposed framework and its components.

\end{itemize}

\section{Related Work}

\subsection{Industrial Anomaly Detection}

Industrial anomaly detection (IAD) has traditionally been studied under two dominant paradigms: \textit{embedding-based} and \textit{reconstruction-based} methods. Embedding-based approaches~\cite{bergmann2019mvtec, defard2021padim, pathcore} detect anomalies by measuring feature deviations from prototypes of normal samples, whereas reconstruction-based models~\cite{zavrtanik2021draem, gong2019memorizing, liznerski2020explainable, li2021cutpaste} learn the distribution of normal data and identify defects through reconstruction residuals. 
However, because they primarily rely on low-level visual discrepancies, their effectiveness often degrades when anomaly recognition requires higher-level semantic reasoning, contextual understanding, or cross-instance comparison.

To address the scarcity of labeled defect samples, subsequent research explored more data-efficient paradigms, including few-shot anomaly detection (FSAD) and zero-shot anomaly detection (ZSAD). Meta-learning approaches~\cite{huang2022registration,wu2021learning} learn transferable representations through large-scale meta-training, while feature-distribution methods such as PatchCore~\cite{pathcore}, SPADE~\cite{spade}, and PaDiM~\cite{defard2021padim} construct compact representations of normal samples to improve data efficiency; more recent works further leverage large-scale pre-trained models for zero-shot anomaly detection. Reconstruction-based approaches built upon masked autoencoders~\cite{schwartz2024maeday} model normal visual patterns to localize anomalies, while CLIP-based methods~\cite{cao2024adaclip,jeong2023winclip,zhou2023anomalyclip} detect anomalies through cross-modal feature alignment. Despite improved generalization ability, these approaches still largely rely on mismatches in pre-trained representations rather than explicitly reasoning about anomaly causes or contextual evidence.

\subsection{MLLM-Based Industrial anomaly detection}

Recent work \cite{wu2023multimodal, yang2025survey} has begun to incorporate Multimodal Large Language Models (MLLMs) into industrial anomaly detection, introducing semantic understanding and reasoning capabilities. Early efforts explored hybrid frameworks that combine language models with specialized vision modules. For example, Myriad \cite{li2023myriad} established a paradigm that couples language reasoning with vision expert models, while AnomalyGPT~\cite{gu2024anomalygpt} enables zero-shot anomaly detection by interpreting feature representations from external detectors. Although effective, these approaches often depend on carefully designed expert models, which may limit scalability across diverse industrial scenarios.

Subsequent studies focus on enhancing reasoning ability within MLLMs. Methods such as LogicAD~\cite{jin2025logicad} and LAD-Reasoner~\cite{li2025lad} introduce chain-of-thought reasoning to improve interpretability, while reinforcement learning–based frameworks, including AnomalyR1~\cite{chao2025anomalyr1}, LR-IAD~\cite{zeng2025lr}, and OmniAD~\cite{zhao2025omniad} optimize multimodal reasoning policies for anomaly detection. However, most existing methods still operate under a single-pass reasoning paradigm without mechanisms to actively acquire additional evidence or verify uncertain predictions. To address this limitation, we propose \textbf{AgentIAD}, an agentic framework that formulates IAD as a closed-loop perception--action process with a unified multi-tool action space for multi-round inspection.

\subsection{Memory-Augmented Agentic Reasoning}

Augmenting agents with external memory has emerged as a promising direction for 
overcoming the limitations of static parametric knowledge. Prior work falls into two 
lines: \textit{visual memory construction}, where agents learn to adaptively request 
finer-grained visual information via reinforcement 
learning~\cite{zheng2025deepeyes, yang2025visionthink, fan2024videoagent}, and \textit{retrieved memory 
construction}, where external knowledge is dynamically retrieved at inference time to 
supplement parametric representations~\cite{lewis2020rag, xu2025amem}. While effective in their respective domains, these two 
directions have been pursued largely in isolation. 

Unlike general visual reasoning, IAD presents a \textbf{dynamic memory challenge}: the evidence required for reliable 
judgment varies substantially across cases, ranging from fine-grained visual details 
in spatially confined defect regions to domain-specific 
normality references beyond the reach of parametric 
knowledge~\cite{gu2024anomalygpt, chao2025anomalyr1}. Visual memory approaches 
improve perceptual granularity but offer no mechanism for external knowledge access; 
retrieval-augmented methods supply such information but are tailored to language-centric 
or video understanding tasks and do not account for the spatially localized nature of 
industrial defects. Neither line of work supports the adaptive, case-by-case memory 
construction that IAD inherently demands.

\section{Method}

\begin{figure*}[t]
    \centering
    \includegraphics[width=1.0\linewidth]{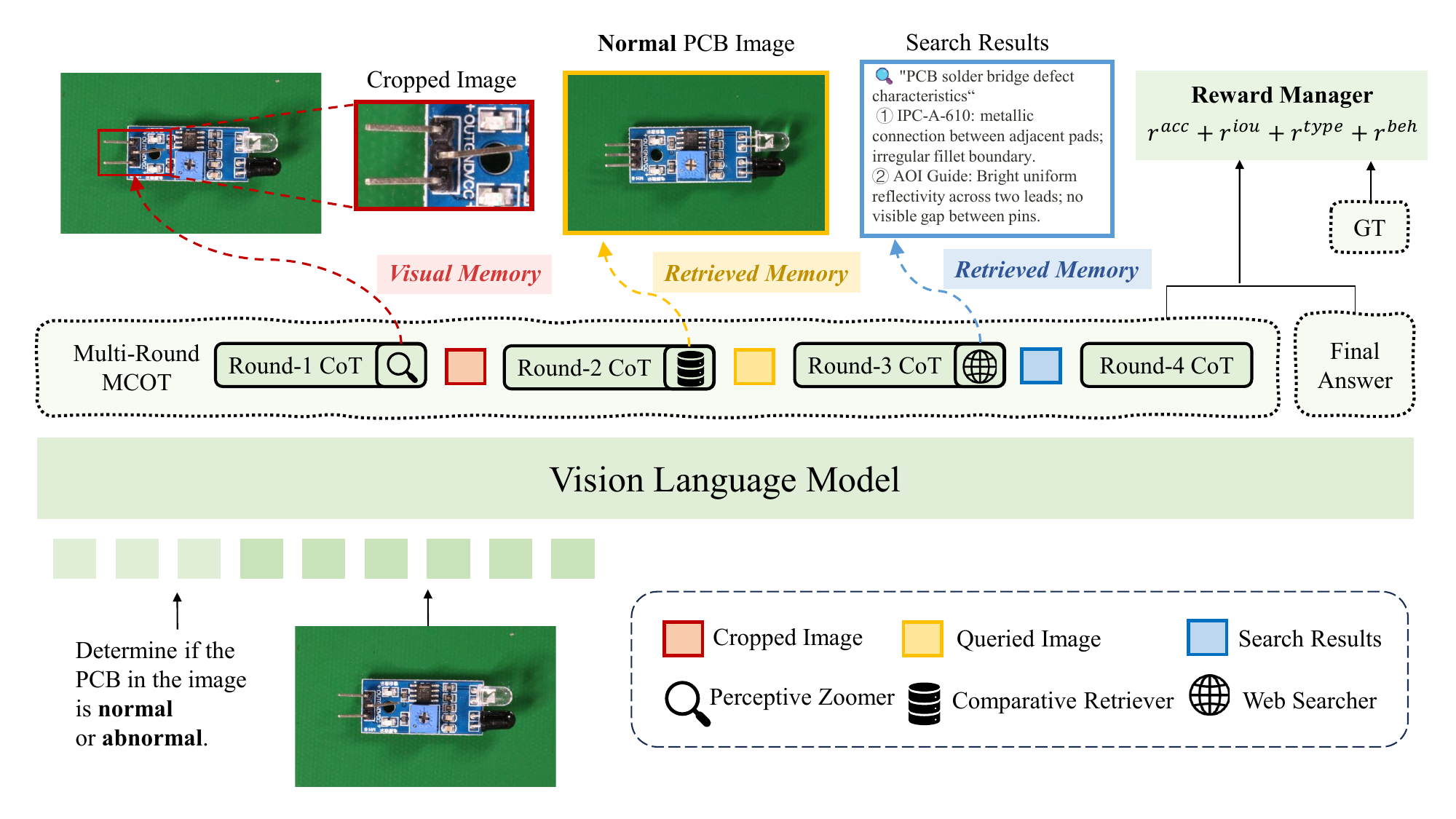}
    \caption{\textbf{Overview of AgentIAD.}
The agent performs iterative industrial inspection through a unified action space with dynamic memory access. 
At each step, it may invoke the \textit{Perceptive Zoomer (PZ)} to retrieve fine-grained visual evidence, the \textit{Comparative Retriever (CR)} for cross-instance verification, or the \textit{Web Searcher (WS)} to access external knowledge, progressively refining its decisions through multi-round perception--action reasoning. 
Training follows a two-stage paradigm: Tool-Aware Supervised Fine-Tuning initializes structured reasoning and tool usage from curated trajectories, followed by Agentic Reinforcement Learning that optimizes long-horizon decision policies under perception--behavior decoupled rewards.
}

    \label{fig:framework}
\end{figure*}

\subsection{Overview}
\label{overview}

We propose \textbf{AgentIAD}, an agentic vision--language framework for industrial anomaly detection that performs iterative inspection through dynamic memory access. The agent operates in a closed-loop perception--action loop and interacts with the environment by invoking memory-access actions. Specifically, the \textit{Perceptive Zoomer (PZ)} retrieves fine-grained \textit{visual memory} through local inspection, while the \textit{Web Searcher (WS)} and \textit{Comparative Retriever (CR)} provide \textit{retrieved memory} via external knowledge acquisition and reference comparison. This design enables progressive evidence accumulation and multi-round reasoning for reliable anomaly analysis. The model is trained using a two-stage strategy that combines Tool-Aware Supervised Fine-Tuning for trajectory initialization with reinforcement learning for policy refinement.

\subsection{Agentic Memory-Augmented Reasoning Framework}

To enable active industrial inspection, we formulate AgentIAD as an agent that performs structured multi-step reasoning through dynamic memory access. Unlike prior IAD methods that rely on passive visual encoding, our framework operates in a closed-loop perception--action manner, allowing the agent to acquire additional evidence by accessing visual and retrieved memory when needed. In practice, memory access is implemented through three tools: the \textbf{Perceptive Zoomer (PZ)}, \textbf{Web Searcher (WS)}, and \textbf{Comparative Retriever (CR)}. Tool selection is embedded directly into the policy space, enabling adaptive multi-stage inspection rather than single-pass prediction.

\subsubsection{Multi-round reasoning trajectory}

We model the industrial inspection process as a Markov Decision Process (MDP). Under this formulation, AgentIAD interacts with the environment in a closed-loop perception--action manner and generates a rollout trajectory through iterative reasoning and memory access.

Given an input image $I$, the agent maintains a state $s_t$ that summarizes the interaction history. At each step $t$, the policy selects an action

\[
a_t \sim \pi_\theta(\cdot \mid s_t),
\]

\noindent where $\pi_\theta$ denotes the agent policy parameterized by $\theta$. If $a_t$ invokes a tool, the environment returns an observation

\[
o_{t+1} = \mathrm{Env}(a_t),
\]

\noindent and the state is updated as

\[
s_{t+1} = s_t \circ (a_t, o_{t+1}).
\]

The interaction continues until the agent emits the \texttt{TERMINATE} action. For interpretability, the policy additionally produces a textual rationale $t_t$ at each step to summarize the current hypothesis and motivate subsequent actions. These rationales serve as explicit reasoning traces while the agent progressively refines its belief by coordinating visual memory access and retrieved memory acquisition.

\subsubsection{Unified memory-access actions}

Within the trajectory, memory access is modeled as a discrete action. AgentIAD operates over a unified action space consisting of:

\begin{itemize}

\item \textbf{Perceptive Zoomer (PZ):} retrieves \textbf{visual memory} through fine-grained local inspection, enabling the model to recover subtle defect cues that may be missed in a global view.

\item \textbf{Web Searcher (WS):} retrieves \textbf{external knowledge} for unfamiliar or rare anomaly types.

\item \textbf{Comparative Retriever (CR):} retrieves \textbf{reference samples} for cross-instance verification.

\end{itemize}

Together, WS and CR provide access to retrieved memory, while PZ provides visual memory. Importantly, these memory-access actions are invoked adaptively by the policy rather than executed in a fixed pipeline. Depending on the inspection context, the agent may selectively access visual memory, retrieved memory, or both, allowing flexible evidence acquisition on a case-by-case basis. This unified action space enables the agent to dynamically compose inspection strategies within a single reasoning trajectory.

\subsection{Two-Stage Training Strategy}

Training an agent capable of multi-round reasoning and memory access is challenging due to the large action space induced by tool interactions. To address this issue, we adopt a pragmatic two-stage training strategy that combines supervised initialization with reinforcement learning.

\subsubsection{Tool-Aware Supervised Fine-Tuning}

\paragraph{Trajectory construction.} We construct structured reasoning trajectories on the MMAD~\cite{jiang2024mmad} dataset with assistance from GPT-4o~\cite{hurst2024gpt}, ensuring coherent visual--language supervision with controlled tool invocations. For each image, a region of interest is obtained from defect bboxes (anomalous samples) or predicted inspection areas (normal samples). The resulting trajectories interleave reasoning traces with staged tool calls, exposing the model to diverse multi-tool interaction patterns. 


\paragraph{Supervised fine-tuning.}

We fine-tune the base VLM on these trajectories using standard supervised learning. At each step, the model autoregressively predicts either a reasoning token or the next tool action, aligning its internal reasoning state with external tool feedback. Unlike conventional instruction tuning that supervises only the final response, our training provides dense intermediate supervision over the perception--action sequence and encourages the model to learn procedural inspection knowledge.

To stabilize training and prevent overfitting to exploratory reasoning noise, we apply a selective loss mask that focuses supervision on decisive steps, reducing the impact of intermediate reasoning errors or incorrect tool invocations. Formally, given a trajectory token sequence
\[
y = \{y_1, \ldots, y_T\}
\]
with mask
\[
M = \{m_1, \ldots, m_T\}, \quad m_t \in \{0,1\},
\]
the supervised objective is
\[
\mathcal{L}_{\text{SFT}}
=
-\sum_{t=1}^{T} m_t \log p_\theta(y_t \mid y_{<t}, x),
\]
where $m_t = 1$ only for the final reasoning response and the last tool invocation. This masking strategy focuses optimization on conclusive reasoning and stabilizes action--decision consistency, providing a strong initialization for subsequent policy optimization.

\subsubsection{Agentic Reinforcement Learning}

While supervised fine-tuning learns structured perception--reasoning patterns, the resulting model remains largely passive. We therefore introduce Agentic Reinforcement Learning, where the agent learns adaptive multi-round policies under an enlarged discrete action space and sparse terminal supervision. Built upon GRPO, this stage refines the perception--action policy through interaction with the environment.

\paragraph{Group Relative Policy Optimization.}

We optimize the agent policy $\pi_\theta(a_t \mid s_t)$ under the GRPO framework. Each trajectory $\tau=\{(s_t,a_t,r_t)\}_{t=1}^{T}$ consists of the state $s_t$, the tool action $a_t$, and reward $r_t$. The objective maximizes a clipped surrogate while regularizing the policy toward the SFT reference policy $\pi_{\mathrm{ref}}$:

\begin{equation}
\mathcal{L}_{\text{clip}}
= \mathbb{E}_{t}\!\left[\min\!\big(\rho_t A_t,\; \tilde{\rho}_t A_t\big)\right],
\end{equation}

\begin{equation}
\mathcal{L}_{\text{KL}}
= D_{\mathrm{KL}}\!\big(\pi_{\theta}\,\|\,\pi_{\mathrm{ref}}\big),
\end{equation}

\begin{equation}
\mathcal{L}_{\text{GRPO}}
= \mathcal{L}_{\text{clip}} - \beta\,\mathcal{L}_{\text{KL}},
\end{equation}

\noindent where
\[
\rho_t=\frac{\pi_{\theta}(a_t \mid s_t)}{\pi_{\mathrm{ref}}(a_t \mid s_t)},
\quad
\tilde{\rho}_t=\operatorname{clip}(\rho_t,1-\epsilon,1+\epsilon),
\]
and $A_t$ denotes the advantage estimated via GAE. Expectations are taken over rollout steps and batch samples.

\paragraph{Reward Overview.}
\label{subsec:reward}

To guide both perception quality and action strategy, we design a two-part reward:

\begin{equation}
R =
\alpha R_{\text{perc}} +
\beta R_{\text{beh}},
\qquad
\alpha,\beta > 0,
\end{equation}

\noindent where $R_{\text{perc}}$ evaluates \emph{what} the agent perceives and understands, while $R_{\text{beh}}$ regulates \emph{how} the agent acts. This decomposition follows the dual objectives of industrial inspection: perceptual correctness and disciplined decision behavior.

\paragraph{Perception Reward.}

The perception reward supervises factual, spatial, and semantic correctness:

\begin{equation}
R_{\text{perc}}
=
R_{\text{acc}} + R_{\text{iou}} + R_{\text{type}} .
\end{equation}

\paragraph{Accuracy Reward.}
\begin{equation}
R_{\text{acc}} =
\mathbb{I}\!\left[\text{format}(\hat{y}_{K})=\text{valid}\right]
\cdot
\mathbb{I}\!\left[\hat{y}_{K}=y_{\text{gt}}\right].
\end{equation}

\paragraph{IoU Reward.}
\begin{equation}
R_{\text{iou}}=
\begin{cases}
1, & \text{if }\mathrm{IoU}(b_{\text{pred}}, b_{\text{gt}}) > 0.5, \\
\mathrm{IoU}(b_{\text{pred}}, b_{\text{gt}}), & \text{otherwise}.
\end{cases}
\end{equation}

\paragraph{Type Reward.}
\begin{equation}
R_{\text{type}} =
\lambda_{\text{type}}
\mathbb{I}[y_{\text{gt}}=1]
\mathbb{I}\!\left[\hat{c}_{K}=c_{\text{gt}}\right].
\end{equation}

\paragraph{Behavior Reward.}

While $R_{\text{perc}}$ focuses on perceptual correctness, the behavior reward $R_{\text{beh}}$ regulates tool-use discipline and efficiency.

\begin{equation}
R_{\text{beh}}
=\frac{1}{K}\!\sum_{t=1}^{K}
\!\left[
\lambda_1\mathbb{I}(\hat{y}_t=y_{\text{gt}})
+\lambda_2 q_t
-\lambda_3\!\max(0,n_t-\bar{n})
\right],
\end{equation}

\noindent where $n_t$ is the number of tool invocations at step $t$, and $q_t$ denotes the normalized frequency of using the tools within the rollout group. This formulation encourages accurate decisions with minimal yet purposeful tool use.

\section{Experiments}
\subsection{Experiment Settings}

\paragraph{Datasets.}
We evaluate our method on industrial anomaly detection benchmark and a medical imaging dataset.
For industrial anomaly detection, we use MMAD~\cite{jiang2024mmad} as the primary benchmark. MMAD evaluates multiple reasoning-based subtasks, including Anomaly Discrimination, Defect Classification, Defect Localization, Defect Description, Defect Analysis, Object Classification, and Object Analysis.
The benchmark integrates four industrial anomaly detection datasets: MVTec-AD~\cite{bergmann2019mvtec}, VisA~\cite{zou2022spot}, MVTec-LOCO~\cite{bergmann2022beyond}, and GoodsAD~\cite{zhang2024pku}. MVTec-AD and VisA mainly contain texture-level defects, while MVTec-LOCO and GoodsAD focus on structural and logical anomalies. Following prior works such as OmniAD~\cite{zhao2025omniad} and AD-FM~\cite{liao2025ad}, we adopt the same data split protocol of MMAD, where 1,936 samples are used for training and 6,400 samples are used for evaluation. For medical anomaly detection, we use the Brain Tumor MRI Dataset~\cite{msoud_nickparvar_2026}. This dataset contains brain MRI scans with tumor labels. Following the original dataset split, we sample 1,900 images from the training set for training and the original test set is used for evaluation.

\paragraph{Evaluation.}
To evaluate our model, we follow the MMAD benchmark protocol and report accuracy for MMAD question answering. For anomaly detection on both industrial and medical datasets, we report classification accuracy, F1-score and image-level AUROC. For MMAD reasoning evaluation, we follow the original question–answering format defined in the benchmark. 
For anomaly detection experiments and ablation studies, we adopt a prompt format similar to PromptAD~\cite{li2024promptad}. 

\paragraph{Implementation Details.}
AgentIAD is built upon Qwen2.5-VL-3B and trained on 8 A100 (80GB) GPUs. 
During SFT, the vision encoder is frozen and the remaining components are optimized using AdamW with cosine learning rate decay. The initial learning rate is set to $2\times10^{-5}$ and training runs for 20 epochs.
During reinforcement learning training, we adopt the OpenRLHF framework and generate 8 rollouts for each prompt to improve exploration during training. The learning rate is set to $1\times10^{-6}$ and training runs for 3 epochs.

\subsection{Main Results}

\begin{table*}[t]
\centering
\caption{Performance comparison of both proprietary and open-source MLLMs in MMAD. All methods we compare are under the standard 1-shot in-context learning setting. Anomaly Discrimination uses the average accuracy of normal and abnormal categories.}
\label{tab:mmad_reasoning}


\resizebox{\textwidth}{!}{
\begin{tabular}{lccccccccc}
\toprule
\textbf{Model} & \textbf{Scale} &
\multicolumn{1}{c}{\textbf{Anomaly}} &
\multicolumn{4}{c}{\textbf{Defect}} &
\multicolumn{2}{c}{\textbf{Object}} &
\textbf{Average} \\

\cmidrule(lr){3-3}
\cmidrule(lr){4-7}
\cmidrule(lr){8-9}

& &
\textbf{Discrimination} &
\textbf{Classification} &
\textbf{Localization} &
\textbf{Description} &
\textbf{Analysis} &
\textbf{Classification} &
\textbf{Analysis} & \\

\midrule

Random Chance & -- & 50.00 & 25.00 & 25.00 & 25.00 & 25.00 & 25.00 & 25.00 & 28.57 \\

Human (expert)~\cite{jiang2024mmad} & -- & 95.24 & 75.00 & 92.31 & 83.33 & 94.20 & 86.11 & 80.37 & 86.65 \\
Human (ordinary)~\cite{jiang2024mmad} & -- & 86.90 & 66.25 & 85.58 & 71.25 & 81.52 & 89.58 & 69.72 & 78.69 \\

\midrule

Claude-3.5-sonnet & -- & 60.14 & 60.14 & 48.81 & 67.13 & 79.11 & 85.19 & 79.83 & 68.36 \\
Gemini-1.5-flash~\cite{team2024gemini} & -- & 58.58 & 54.70 & 49.10 & 66.53 & 82.24 & 91.47 & 79.71 & 68.90 \\
Gemini-1.5-pro~\cite{team2024gemini} & -- & \textbf{68.63} & 60.12 & \textbf{58.56} & 70.38 & 82.46 & 89.20 & 82.25 & 73.09 \\
GPT-4o-mini & -- & 64.33 & 48.58 & 38.75 & 63.68 & 80.40 & 88.56 & 79.74 & 66.29 \\
GPT-4o & -- & \textbf{68.63} & \textbf{65.80} & 55.62 &\textbf{ 73.21} & \textbf{83.41} & \textbf{94.98} & \textbf{82.80} &\textbf{ 74.92} \\

\midrule

AnomalyGPT~\cite{gu2024anomalygpt} & 7B & 65.57 & 27.49 & 27.97 & 36.86 & 32.11 & 29.84 & 35.82 & 36.52 \\
Qwen-VL-Chat~\cite{bai2023qwen} & 7B & 53.65 & 31.33 & 28.62 & 41.66 & 63.99 & 74.46 & 67.94 & 51.66 \\
LLaVA-1.5~\cite{liu2024improved} & 7B & 51.33 & 37.04 & 36.62 & 50.60 & 69.79 & 68.29 & 69.53 & 54.74 \\
MiniCPM-V2.6~\cite{yao2024minicpm} & 8B & 57.31 & 49.22 & 43.28 & 65.86 & 75.24 & 92.02 & 80.80 & 66.25 \\
InternVL2~\cite{chen2024internvl} & 8B & 59.97 & 43.85 & 47.91 & 57.60 & 78.10 & 74.18 & 80.37 & 63.14 \\
LLaVA-1.5~\cite{liu2023visual} & 13B & 49.96 & 38.78 & 46.17 & 58.17 & 73.09 & 73.62 & 70.98 & 58.68 \\
LLaVA-NeXT~\cite{liu2024llavanext} & 34B & 57.92 & 48.79 & 52.87 & 71.34 & 80.28 & 81.12 & 77.80 & 67.16 \\
InternVL2~\cite{chen2024internvl} & 76B & 68.25 & 54.22 & 56.66 & 66.30 & 80.47 & 86.40 & 82.92 & 70.75 \\
Qwen-2.5-VL~\cite{bai2025qwen3} & 3B & 51.10 & 41.07 & 44.87 & 61.43 & 75.99 & 86.54 & 79.58 & 62.94 \\
AnomalyR1~\cite{chao2025anomalyr1} & 3B & 60.62 & 63.56 & 70.14 & 80.47 & 85.28 & 92.48 & 86.15 & 76.96 \\
\textbf{AgentIAD (ours)} & \textbf{3B} & \textbf{69.49} & \textbf{72.73} & \textbf{80.94} & \textbf{85.27} & \textbf{87.84} & \textbf{93.27} & \textbf{90.59} & \textbf{82.88} \\

\bottomrule
\end{tabular}
}
\end{table*}

\paragraph{Results on MMAD benchmark.}
Table~\ref{tab:mmad_reasoning} presents that AgentIAD achieves high score on the MMAD benchmark particularly in defect-related reasoning tasks. The improvement is most significant in Defect Localization, where AgentIAD substantially outperforms other MLLMs. AgentIAD also performs competitively on object-level reasoning tasks and obtains the highest score on Object Analysis.

These results demonstrate that adaptive memory augmentation is critical for anomaly understanding. Specifically, the Perceptive Zoomer (PZ) constructs fine-grained visual memory to mitigate the visual memory deficiency of static models, while the Comparative Retriever (CR) and Web Searcher (WS) provide retrieved memory to address retrieved memory deficiency in complex analysis. Under our agentic reinforcement learning framework, these memory-augmentation tools are invoked adaptively, allowing the model to better align its perception and reasoning with the requirements of diverse subtasks.

\begin{table}[t]
\centering
\caption{Performance comparison on the Brain Tumor MRI dataset~\cite{msoud_nickparvar_2026}.
We compare the no-tool baseline GPT-4o with AgentIAD models trained with different reasoning tools.}
\label{tab:brain_results}

\setlength{\tabcolsep}{6pt}
\renewcommand{\arraystretch}{1.1}
\scriptsize

\begin{tabular}{lcccc}
\toprule
\textbf{Model} & \textbf{Tool} & \textbf{Accuracy} & \textbf{F1-score} & \textbf{AUROC} \\
\midrule
GPT-4o & -- & 79.31 & 87.80 & 59.38 \\
AgentIAD & PZ & 91.06 & 94.14 & 86.38 \\
AgentIAD & PZ + WS & \textbf{95.50} & \textbf{97.00} & \textbf{93.83} \\
\bottomrule
\end{tabular}

\end{table}

\paragraph{Results on Medical benchmark with Web Searcher.}

As shown in Table~\ref{tab:brain_results}, we further study the impact of the WS tool with the Brain Tumor MRI dataset~\cite{msoud_nickparvar_2026}. 
During training, we only sample 1,900 images from the 5,600 training samples. We train two variants of AgentIAD: one using only the PZ tool and another using both PZ and WS.
Compared to the model using only Visual Memory (via PZ), the model incorporating Retrieved Memory (via WS) improves accuracy by 4.44\% and AUROC by 7.45\%.
These results demonstrate that while Visual Memory is essential for inspecting suspicious regions, Retrieved Memory provides critical semantic knowledge to interpret complex visual patterns.
The synergy between these two memory types enables the agent to overcome both visual and retrieved memory deficiencies, leading to more reliable anomaly analysis.

\paragraph{Anomaly Detection Results}
We evaluate the anomaly detection capability of AgentIAD after the GRPO training stage. 
Table~\ref{tab:auroc_results} compares AgentIAD with several methods on the MVTec and VisA benchmarks .
For training, we follow the dataset split protocol introduced in OmniAD~\cite{zhao2025omniad}. 
Following PromptAD\cite{li2024promptad}, we adopt a similar prompt formulation during inference and evaluate the image-level AUROC. To ensure a fair comparison, we remove from the PromptAD test set any samples that overlap with our training set.
During inference, reference images are only introduced when the agent invokes the Comparative Retriever (CR) tool to retrieve one normal example for comparison. Otherwise, the model performs anomaly detection directly without a reference image. All other compared methods are evaluated under a strict 1-shot setting. As shown in Table~\ref{tab:auroc_results}, AgentIAD achieves strong performance on both datasets. 
These results indicate that adaptive memory augmentation is highly effective for anomaly detection in few-shot settings. By dynamically building visual and retrieved memory, the agent can achieve more reliable and accurate detection.

\begin{table}[h]
\centering

\begin{minipage}[t]{0.32\linewidth}
\centering
\caption{Comparison of image-level AUROC on MVTec and VisA datasets.}
\label{tab:auroc_results}

\tiny
\setlength{\tabcolsep}{3pt}
\renewcommand{\arraystretch}{1.05}

\begin{tabular}{lcc}
\toprule
\textbf{Method} & \textbf{MVTec} & \textbf{VisA} \\
\midrule
SPADE\cite{cohen2020sub} & 81.0 & 79.5 \\
PaDiM\cite{defard2021padim} & 76.6 & 62.8 \\
PatchCore\cite{roth2022towards} & 83.4 & 79.9 \\
WinCLIP+\cite{jeong2023winclip} & 93.1 & 83.8 \\
RWDAT\cite{tamura2023random} & 93.3 & 83.4 \\
PromptAD\cite{li2024promptad} & 94.6 & 86.9 \\
\midrule
AgentIAD & \textbf{95.3} & \textbf{98.7} \\
\bottomrule
\end{tabular}

\end{minipage}
\hfill
\begin{minipage}[t]{0.65\linewidth}
\centering
\caption{Ablation on reasoning tools and training strategies. All models are based on Qwen2.5-VL-3B, where the middle three variants(+PZ, +CR, and +PZ \& CR) are trained with Tool-Aware SFT.}
\label{tab:ablation_strategy}

\tiny
\setlength{\tabcolsep}{4pt}
\renewcommand{\arraystretch}{0.92}

\begin{tabular}{lccccc}
\toprule
\textbf{Model} & \textbf{MVTec} & \textbf{VisA} & \textbf{LOCO} & \textbf{GoodsAD} & \textbf{Avg.} \\
\midrule
\multicolumn{6}{l}{\textbf{Qwen2.5-VL-3B-Instruct}} \\
\midrule
+ CoT  & 49.73 & 54.93 & 39.27 & 46.56 & 47.62 \\
+ PZ & 95.14 & 93.51 & 89.86 & 90.27 & 92.20 \\
+ CR & 97.22 & \textbf{98.85} & 88.07 & \textbf{97.94} & 95.52 \\
+ PZ \& CR  & 96.91 & 97.27 & 92.50 & 96.54 & 95.81 \\
+ \textbf{AgentIAD}  & \textbf{97.84} & 98.54 & \textbf{95.57} & 97.90 & \textbf{97.46} \\
\bottomrule
\end{tabular}

\end{minipage}

\end{table}

\subsection{Ablation Studies}

In all ablation experiments, we adopt a prompt format similar to PromptAD~\cite{li2024promptad}, and the training and test splits follow the protocol introduced in OmniAD~\cite{zhao2025omniad}.

\paragraph{Effect of tool usage and training strategies.}
Table~\ref{tab:ablation_strategy} analyzes the contribution of the proposed memory modules and training stages.
Starting from the CoT-only baseline, the model shows limited anomaly detection capability. Introducing Visual Memory (via Perceptive Zoomer) leads to a large improvement, confirming that inspecting localized regions is critical for identifying subtle anomalies.
Similarly, Retrieved Memory (via Comparison Retriever) significantly boosts performance by allowing the model to verify samples against normal references.
Combining both memory types under SFT yields further gains, suggesting that localized inspection and reference comparison provide complementary signals.
Finally, applying GRPO to jointly optimize memory access results in the AgentIAD model, which achieves the best performance across all datasets.

\begin{table}[h]
\centering
\caption{\textbf{Ablation on reward components.} Binary classification accuracy (\%) across four benchmarks and corresponding tool usage statistics.}
\label{tab:reward_ablation}

\setlength{\tabcolsep}{4pt}
\renewcommand{\arraystretch}{0.92}
\footnotesize

\resizebox{\textwidth}{!}{
\begin{tabular}{>{\centering\arraybackslash}m{3cm}cccccccc}
\toprule
 & \multicolumn{5}{c}{\textbf{Accuracy}} & \multicolumn{3}{c}{\textbf{Tools}} \\
\cmidrule(lr){2-6} \cmidrule(lr){7-9}
\textbf{Model} & \textbf{MVTec} & \textbf{VisA} & \textbf{LOCO} & \textbf{GoodsAD} & \textbf{Avg} & \textbf{\#PZ} & \textbf{\#CR} & \textbf{Total} \\
\midrule
\textbf{AgentIAD} & 97.84 & \textbf{98.54} & \textbf{95.57} & \textbf{97.90} & \textbf{97.46} & 6400 & 374 & 6774 \\
w/o $R_{\text{beh}}$ & \textbf{97.92} & 97.76 & 93.87 & 97.54 & 96.77 & 6401 & 298 & 6699 \\
w/o $R_{\text{perc}}$ + $R_{\text{beh}}$ & 96.91 & 97.27 & 92.50 & 96.54 & 95.81 & 6397 & 642 & 7039 \\
\bottomrule
\end{tabular}
}

\end{table}

\paragraph{Effect of reward components.}
Table~\ref{tab:reward_ablation} analyzes the influence of different reward components on both task accuracy and tool usage behavior.
Removing both the perception reward $R_{\text{perc}}$ and the behavior reward $R_{\text{beh}}$ leads to a noticeable decrease in overall performance. In addition, the number of tool invocations increases substantially, indicating that without reward guidance the agent tends to rely on tools more frequently without necessarily improving decision quality.
When only the behavior reward $R_{\text{beh}}$ is removed, the overall accuracy also decreases slightly. Compared with the full model, the usage of the CR tool becomes less balanced, suggesting that $R_{\text{beh}}$ helps regulate how the agent invokes different tools during reasoning.
The full reward formulation achieves the best overall accuracy while maintaining a more controlled tool usage pattern. These observations suggest that $R_{\text{perc}}$ mainly supports perceptual correctness, while $R_{\text{beh}}$ helps shape more disciplined tool interaction during reasoning.

\paragraph{Effect of tool design and training.}
We first analyze the distribution of anomaly area ratios in MMAD. 
As shown in Tab.~\ref{tab:mmad_area}, anomalies in MMAD are highly localized,
with a median area ratio of only 0.58\%.
This observation motivates the design of the Perceptive Zoomer (PZ) tool,
 which enables the agent to construct fine-grained visual memory by zooming into small anomalous regions. We further evaluate the importance of training for tool usage. 
Tab.~\ref{tab:gpt4o_tools} compares GPT-4o equipped with the same PZ and CR tools with our AgentIAD.  
The result indicates that tool-aware training is crucial for enabling the agent to effectively coordinate adaptive memory augmentation during anomaly reasoning.

\begin{table}[h]
\centering
\begin{minipage}[c]{0.45\linewidth}
\centering
\caption{Distribution of anomaly area ratios in the MMAD. The statistics are computed from abnormal samples.}
\label{tab:mmad_area}

\tiny
\setlength{\tabcolsep}{2.5pt}
\renewcommand{\arraystretch}{0.9}

\begin{tabular}{ccccccc}
\toprule
Mean & Median & P75 & P90 & P95 & P99 & Max \\
\midrule
3.21\% & 0.58\% & 2.78\% & 8.70\% & 14.54\% & 34.82\% & 100\% \\
\bottomrule
\end{tabular}

\end{minipage}
\hfill
\begin{minipage}[c]{0.5\linewidth}
\centering
\caption{Accuracy comparison between GPT-4o with PZ/CR tools and AgentIAD.}
\label{tab:gpt4o_tools}

\tiny
\setlength{\tabcolsep}{3pt}
\renewcommand{\arraystretch}{0.9}

\begin{tabular}{lccccc}
\toprule
Model & MVTec & VisA & LOCO & GoodsAD & Avg. \\
\midrule
GPT-4o w. tools & 85.57 & 77.20 & 45.91 & 72.56 & 70.31 \\
AgentIAD & 97.84 & 98.54 & 95.57 & 97.90 & 97.46 \\
\bottomrule
\end{tabular}

\end{minipage}

\end{table}

\subsection{Visualization Results}

\begin{figure*}[t]
    \centering
    \includegraphics[width=1\linewidth]{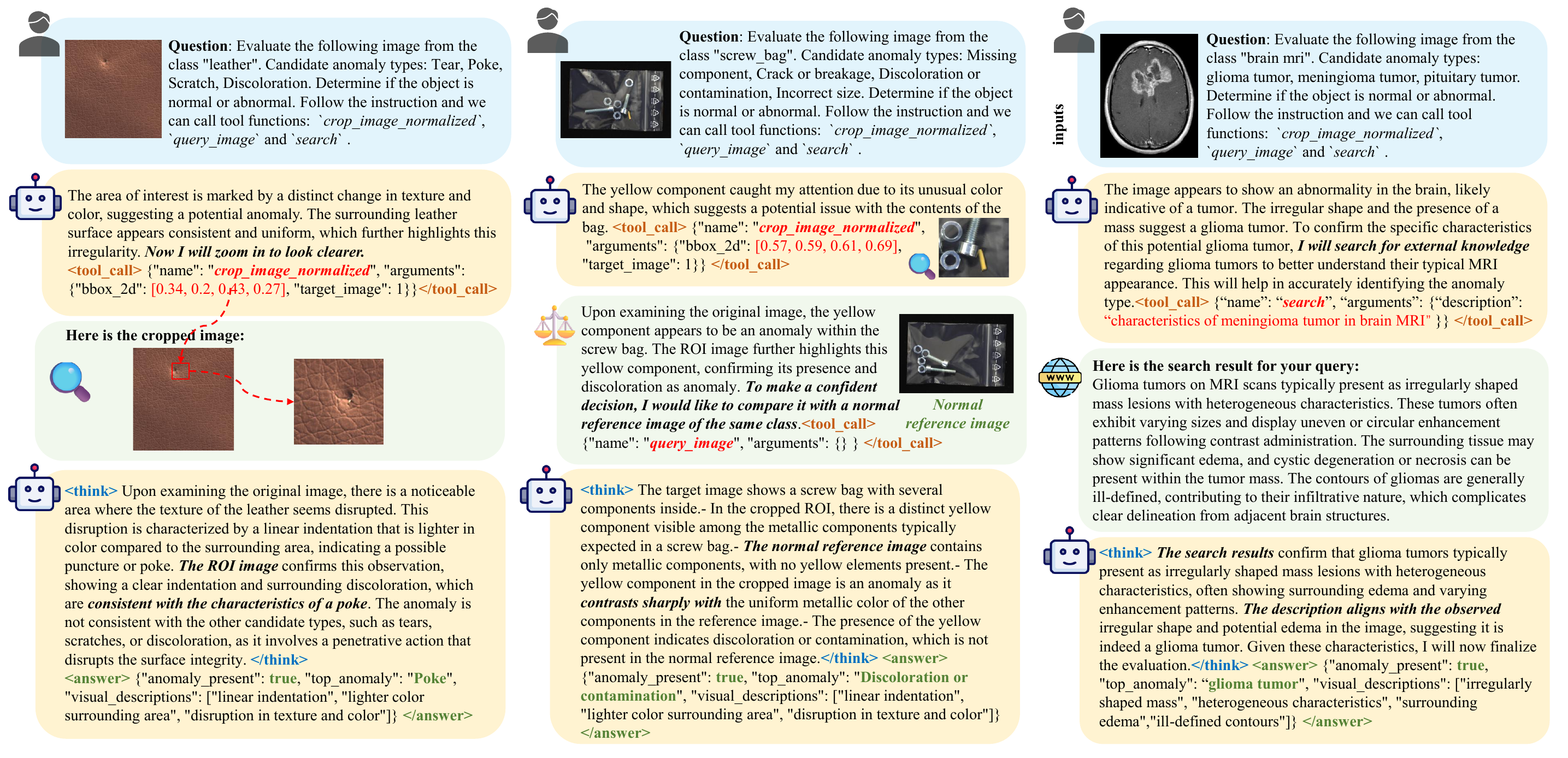}
    \caption{\textbf{Visualize Inference Cases of AgentIAD.}Three inference cases illustrating how AgentIAD adaptively constructs \textbf{Visual Memory} (left) and \textbf{Retrieved Memory} (middle \& right) to support anomaly reasoning across various scenarios.}

    \label{fig:vis_result}
\end{figure*}

Figure~\ref{fig:vis_result} presents three inference cases illustrating how AgentIAD adaptively manages its memory during reasoning.
The left example shows the construction of Visual Memory via the \textit{Perceptive Zoomer}. The agent adaptively zooms into a suspicious region to recover fine-grained details, effectively resolving the visual memory deficiency of static models.
The middle and right examples demonstrate the acquisition of Retrieved Memory. In the middle case, the agent invokes the \textit{Comparative Retriever} to retrieve a normal reference, using episodic memory to verify ambiguous cues. In the right case, the agent utilizes the \textit{Web Searcher} to access semantic knowledge in the medical domain, helping it interpret complex visual patterns.
These examples highlight how AgentIAD overcomes retrieved memory deficiency by flexibly querying external knowledge sources to support reliable anomaly analysis.

\section{Conclusion}

In this work, we presented \textbf{AgentIAD}, an agentic framework for industrial anomaly detection that performs iterative inspection through a perception--action loop. By enabling dynamic access to both \emph{visual memory} and \emph{retrieved memory}, AgentIAD progressively acquires complementary evidence and refines its decisions during inference. To train such an agent effectively, we adopt a two-stage strategy that combines tool-aware supervised fine-tuning with agentic reinforcement learning, enabling stable learning of long-horizon perception--action policies. Extensive experiments demonstrate that AgentIAD achieves state-of-the-art performance on the MMAD benchmark while producing interpretable inspection traces that closely resemble human quality inspection procedures. We hope this work highlights the potential of memory-augmented agentic reasoning for reliable and interpretable visual inspection.

\noindent \textbf{Limitation.} Despite these promising results, our current implementation is built upon Qwen2.5-VL-3B rather than the latest multimodal architectures such as Qwen3.5. Future work could integrate more advanced vision–language models and expand the set of inspection tools to further improve generalization, robustness, and cross-domain adaptability.




%
%
\bibliographystyle{splncs04}
\bibliography{main}
\clearpage
\section{Supplementary Material}

\subsection{Tool Implementation Details}

\subsubsection{Perceptive Zoomer}

The Perceptive Zoomer (PZ) enables the agent to access visual memory by inspecting local regions of the image.

Given a bounding box predicted by the agent in normalized coordinates $bbox = (x_1, y_1, x_2, y_2)$, the environment crops the corresponding region from the original image and returns the cropped image as a new visual observation.

Bounding boxes are predicted by the model through the following structured tool call:

\begin{verbatim}
<tool_call>{
"name": "crop_image_normalized",
"arguments": {"bbox_2d": [...], "target_image": 1}
}</tool_call>
\end{verbatim}

The cropped image is appended to the interaction history, allowing the agent to refine its reasoning with finer visual details.

\subsubsection{Comparative Retriever}

The Comparative Retriever (CR) enables the agent to access retrieved memory by comparing the current image with a normal reference instance during anomaly reasoning.

For samples from the MMAD benchmark, the dataset provides
reference images through the \texttt{similar\_templates}
field. When the CR tool is invoked, the system returns a
single normal image from this field.

For the Brain Tumor MRI dataset, retrieved memory is constructed from normal samples in the training split. Image similarity is measured using CLIP embeddings (ViT-B-32), and for each sample the most similar normal training image is identified.

During inference, the system directly returns this
pre-identified normal reference image without performing
additional similarity search.

The CR tool is invoked through the following structured function call:

\begin{verbatim}
<tool_call>{
"name":"query_image",
"arguments":{}
}</tool_call>
\end{verbatim}

After execution, the environment appends the retrieved reference image to the interaction history as an additional visual observation.

\subsubsection{Web Searcher}

The Web Searcher (WS) enables the agent to access retrieved memory in the form of external textual knowledge.

When the WS tool is invoked, the agent provides a natural language description as the search query. The system retrieves relevant information using the web search service provided by Zhipu AI. 

The resulting description is returned as textual context and appended to the interaction history, allowing the agent to incorporate external knowledge during reasoning.

The WS tool is invoked through the following
structured function call:

\begin{verbatim}
<tool_call>{
"name":"search",
"arguments":{"description":"..."}
}</tool_call>
\end{verbatim}

WS is used only to retrieve general domain knowledge rather than dataset-specific information.

\subsection{Bounding Box Usage in Training}
Bounding box annotations are used only during the trajectory
construction stage of supervised fine-tuning (SFT).

These bounding boxes are used to construct demonstration
trajectories that guide the Perceptive Zoomer toward plausible
inspection regions.

For samples with pixel-level anomaly masks, bounding boxes are
obtained by computing the bounding box of the mask region. For
samples without mask annotations, approximate inspection regions
are used during trajectory construction to simulate the region that
a human inspector might examine.

Bounding boxes are not provided as inputs to the model.
During reinforcement learning and inference, the agent predicts
bounding boxes autonomously through tool calls. For the perception
reward used in reinforcement learning, the reference bounding boxes
used for IoU computation follow the same sources described above
(i.e., mask-derived boxes or approximate inspection regions) and
are used only for reward evaluation rather than as model inputs.

\subsection{Notation Clarification}

For completeness, we clarify several symbols used in the reinforcement learning formulation in Sec.~3.3 of the main paper.

\subsubsection{Trajectory length.}
The symbol $K$ denotes the total number of reasoning steps in a rollout trajectory. The final prediction is produced at step $K$, and $\hat{y}_K$ therefore represents the agent’s final anomaly prediction.

\subsubsection{Prediction variables.}
The variable $\hat{y}_t$ denotes the prediction produced by the agent at reasoning step $t$, while $y_{\text{gt}}$ denotes the corresponding ground-truth anomaly label.  
Similarly, $\hat{c}_K$ and $c_{\text{gt}}$ denote the predicted and ground-truth anomaly categories at the final reasoning step.

\subsubsection{Behavior reward variables.}
In Eq.~(9), $n_t$ denotes the number of tool invocations at reasoning step $t$.  
The quantity $\bar{n}$ represents a reference tool-usage budget used in the reward formulation.  
The variable $q_t$ denotes a normalized score reflecting the frequency of tool usage at step $t$ within the rollout group during training.

\subsubsection{Policy optimization variables.}
In the GRPO objective, $\rho_t = \frac{\pi_\theta(a_t \mid s_t)}{\pi_{\mathrm{ref}}(a_t \mid s_t)}$ denotes the policy ratio between the current policy and the reference policy.  
The clipped ratio $\tilde{\rho}_t = \operatorname{clip}(\rho_t, 1-\epsilon, 1+\epsilon)$ uses $\epsilon$ as the clipping threshold, and $A_t$ denotes the advantage estimate used for policy optimization.

\subsection{Additional Training Details}
We adopt the GRPO framework to the SFT reference policy in the reinforcement learning stage.
For completeness, the main optimization hyperparameters are summarized in Table~\ref{tab:grpo_hparams}, and the reward coefficients are listed in Table~\ref{tab:reward_coeffs}.

\begin{table}[t]
\centering

\begin{minipage}[t]{0.48\linewidth}
\centering
\caption{Main hyperparameters used in GRPO training.}
\label{tab:grpo_hparams}
\small
\begin{tabular}{lc}
\toprule
\textbf{Hyperparameter} & \textbf{Value} \\
\midrule
Optimizer & AdamW \\
Learning rate & $1\times10^{-6}$ \\
Rollouts per prompt & 8 \\
Training epochs & 3 \\
Clipping ratio $\epsilon$ & 0.2 \\
Temperature & 1.0 \\
Precision & bfloat16 \\
\bottomrule
\end{tabular}
\end{minipage}
\hfill
\begin{minipage}[t]{0.48\linewidth}
\centering
\caption{Reward coefficients used in GRPO training.}
\label{tab:reward_coeffs}
\small
\begin{tabular}{lc}
\toprule
\textbf{Parameter} & \textbf{Value} \\
\midrule
$\alpha$ (perception reward weight) & 1.0 \\
$\beta$ (behavior reward weight) & 1.0 \\
$\lambda_{\text{type}}$ (type reward weight) & 0.1 \\
$\lambda_1$ (correctness term) & 1.0 \\
$\lambda_2$ (tool-usage term) & 0.5 \\
$\lambda_3$ (efficiency penalty) & 0.05 \\
$\bar{n}$ (reference tool budget) & 1.0 \\
\bottomrule
\end{tabular}
\end{minipage}

\end{table}

\begin{table}[t]
\centering
\setlength{\tabcolsep}{10pt}
\caption{Inference latency of the AgentIAD (PZ+WS) model reported in Table~\ref{tab:brain_results} on the Brain Tumor MRI test set. Latency is measured as the average wall-clock time per sample.}
\label{tab:ws_latency}
\small
\begin{tabular}{lccc}
\toprule
\textbf{Setting} & \textbf{Samples} & \textbf{WS Usage (\%)} & \textbf{Latency (s/sample)} \\
\midrule
Overall & 1600 & 18.2 & 2.90 \\
No WS invoked & 1309 & 0.0 & 2.82 \\
WS invoked & 291 & 100.0 & 3.26 \\
\bottomrule
\end{tabular}
\end{table}

\subsection{Inference Cost of the Web Searcher}

We measure the inference cost introduced by the Web Searcher (WS) on the Brain Tumor MRI test set. 
The measurement is conducted using the same AgentIAD (PZ+WS) model reported in Table~\ref{tab:brain_results}. 
Latency is measured as the average wall-clock inference time per sample with batch size 1 under the same hardware and decoding settings.

Table~\ref{tab:ws_latency} reports the latency statistics. 
The average inference time is 2.90 seconds per sample, and WS is invoked in 18.2\% of the cases, indicating that the agent performs web search only when additional information is needed.

When WS is not invoked, the average inference time is 2.82 seconds per sample. 
When WS is triggered, the latency increases to 3.26 seconds per sample, corresponding to an additional overhead of about 0.45 seconds.
Overall, WS introduces a small increase in inference time, while the overall latency remains moderate because the search tool is used selectively.

\end{document}